%% file: gradient_template.tex
\documentclass[]{gradient} % For LaTeX2e

\usepackage{array} 

\usepackage{microtype}
\usepackage{makecell}
\usepackage{textgreek} 
\usepackage{graphicx}

\usepackage{booktabs} 
\usepackage[table,xcdraw]{xcolor}
\usepackage{xurl}

\usepackage{hyperref}

\usepackage{subcaption}

\usepackage{tcolorbox}
\tcbuselibrary{listings}

\usepackage{amsmath}
\usepackage{amssymb}
\usepackage{mathtools}
\usepackage{amsthm}

\usepackage{natbib}

\usepackage{colortbl}   
\usepackage{wrapfig}

\usepackage{mathpazo}

\usepackage[linesnumbered,ruled,vlined]{algorithm2e}
\usepackage{listings}

\usepackage{multirow}
\usepackage{siunitx}
\usepackage{tikz}
\usepackage{enumitem}

\title{EVM-QuestBench: An Execution-Grounded Benchmark for Natural-Language Transaction Code Generation}
\author{
    Pei Yang\textsuperscript{1*}
    \quad Wanyi Chen\textsuperscript{2*}
    \quad Ke Wang\textsuperscript{1}
    \quad Lynn Ai\textsuperscript{1}
    \quad Eric Yang\textsuperscript{1}
    \quad Tianyu Shi\textsuperscript{1\textdagger}
}
 
\affiliation{
    \textsuperscript{1}Gradient \quad
    \textsuperscript{2}Soochow University \\
    \textsuperscript{*}Equal contribution \quad
    \textsuperscript{\textdagger}Corresponding author
}

\date{Feb 26, 2026}
\correspondence{ty.shi@mail.utoronto.ca}
\sourcecode{https://github.com/OpenEdgeHQ/EVM-quest-bench}

\begin{document}

\abstract{Large language models are increasingly applied to various development scenarios. However, in on-chain transaction scenarios, even a minor error can cause irreversible loss for users. Existing evaluations often overlook execution accuracy and safety. We introduce EVM-QuestBench, an execution-grounded benchmark for natural-language transaction-script generation on EVM-compatible chains. The benchmark employs dynamic evaluation: instructions are sampled from template pools, numeric parameters are drawn from predefined intervals, and validators verify outcomes against these instantiated values. EVM-QuestBench contains 107 tasks (62 atomic, 45 composite). Its modular architecture enables rapid task development. The runner executes scripts on a forked EVM chain with snapshot isolation; composite tasks apply step-efficiency decay. We evaluate 20 models with 5 independent rounds each and find large performance gaps, with split scores revealing persistent asymmetry between single-action precision and multi-step workflow completion. }
\maketitle

\input{tex/01_intro}
\input{tex/02_related}
\input{tex/03_benchmark}
\input{tex/04_setup}
\input{tex/05_results}
\input{tex/06_conclusion}

\input{tex/07_limitations}

% References
\bibliography{custom}

% Appendix
\appendix
\onecolumn
\input{tex/A_appendix}

\end{document}

%% file: tex/01_intro.tex
\section{Introduction}
Large language models (LLMs) are increasingly being used to control software and tools \citep{openai2023gpt4,anil2023palm2}. Leveraging LLMs for code generation and blockchain transactions is becoming commonplace, but this introduces significant financial risks. Even a small error, such as an incorrect address, unit, or deadline, can result in irreversible losses.

Benchmarks for code understanding and generation need to cover a broad range of programming tasks and reasoning capabilities \citep{chen2021evaluating,lu2021codexglue}. Many evaluations still rely on lexical overlap metrics such as BLEU or CodeBLEU \citep{papineni2002bleu,ren2020codebleu}. These metrics can reward outputs that appear similar to references but fail to run or fail to meet functional constraints. Benchmarks such as SWE-bench \citep{jimenez2024swebench} focus on real-world software engineering tasks like bug fixing in Python repositories. Due to the difference in test domains, such benchmarks cannot directly reflect a model's ability to execute transactions in Web2/Web3 environments. Blockchain-specific benchmarks such as Solana Bench \citep{solana2025bench} explore the boundaries of LLM capabilities in Web3 but struggle to provide feedback on the accuracy of natural language understanding and the safety of transaction execution.

Transaction script generation presents a distinct set of failure modes. LLMs must first interpret diverse natural language instructions. The generated code must correctly construct calldata, account for chain-specific units and token decimals, adhere to protocol constraints, and manage dependencies across multiple transaction steps. Even minor deviations can lead to transaction reverts, partial execution, or incorrect state transitions. These characteristics make blockchain automation an ideal domain for execution-based evaluation.

We introduce EVM-QuestBench, an execution-grounded benchmark with two splits. \textbf{Atomic tasks} test single-action precision. \textbf{Composite tasks} test multi-step workflows that require planning, prerequisite handling, and parameter propagation. Composite scoring incorporates a step efficiency factor that penalizes unnecessary steps. EVM-QuestBench contains 107 tasks with a maximum total score of 10{,}700. This design significantly simplifies benchmark development and maintenance. Creating an atomic task only requires defining the problem and developing a validator. Creating a new composite task only requires updating a JSON file.

We evaluate 20 models under a unified protocol with 5 independent rounds per model. Results show substantial variance. Split scores reveal a persistent capability asymmetry between single-action precision and workflow completion. Several models achieve strong Composite performance with weaker Atomic scores. Several models fail on Composite workflows despite non-trivial Atomic performance. These patterns motivate split reporting for diagnosis.

Our contributions are as follows:
\begin{itemize}
  \item We release EVM-QuestBench, a benchmark for natural language to transaction script generation on EVM-compatible chains, with Atomic and Composite splits. Its declarative task design keeps evaluation costs low: each task round consumes a small number of tokens, enabling comprehensive model assessment at minimal token cost.
  \item We introduce an atomic/composite benchmark paradigm that significantly reduces development costs, especially when using LLMs to assist development.
  \item We provide an execution protocol with snapshot isolation, a fixed runner interface, and validator-based scoring over receipts and post-state constraints.
  \item We report results on 20 models with 5-round statistical analysis, including standard deviations, observed ranges, and rank-order consistency, that separates single-action precision from multi-step workflow completion.
\end{itemize}

%% file: tex/02_related.tex
\section{Related Work}
\label{sec:related}

\paragraph{Execution-based evaluation for code generation.}
Reference-based code evaluation metrics such as BLEU and CodeBLEU \citep{papineni2002bleu,ren2020codebleu} share a fundamental flaw: they measure surface similarity rather than functional correctness, so code that resembles a reference but fails at runtime can still receive a high score. Execution-based benchmarks address this by running generated programs against tests, including HumanEval and MBPP for function synthesis \citep{chen2021evaluating,austin2021program}, and extensions to library usage and multilingual settings \citep{wang2023odex,khan2024xcodeeval,yan2024codescope}. However, most prior benchmarks focus on stateless, sandboxed functions and do not model shared external state or irreversible actions.

\paragraph{Real-world software engineering, agent benchmarks, and blockchain constraints.}
SWE-bench evaluates issue resolution by applying patches to real repositories \citep{jimenez2024swebench}, and agent-style benchmarks study multi-step tool use in interactive environments \citep{qin2023toollm,liu2023agentbench}. However, these benchmarks do not target the constraints that characterize blockchain interaction: shared mutable state, protocol prerequisites, strict unit and decimal handling, and irreversible revert risk. Prior blockchain benchmarks such as Solana-focused transaction evaluations \citep{solana2025bench} also do not disentangle single-transaction precision from multi-transaction workflow completion under a unified execution and validation interface. EVM-QuestBench fills this gap by evaluating natural-language-to-transaction-script generation on EVM-compatible chains with atomic/composite splits, using declaratively specified tasks and reusable validator components that keep development costs low while preserving execution-grounded evaluation.

%% file: tex/03_benchmark.tex
\section{Benchmark Tasks}
\label{sec:benchmark}

\subsection{Benchmark Overview}
EVM-QuestBench evaluates whether a model can convert a natural language goal into an executable outcome on EVM-compatible chains. Each task specifies a target end state. For evaluation, the runner samples a template from a pre-built pool and instantiates numeric parameters within predefined ranges. The model outputs a TypeScript module that constructs transaction request objects. The runner executes the plan on a forked chain, and validators score post-execution constraints.

\begin{figure*}[t]
  \centering
  \includegraphics[width=\textwidth]{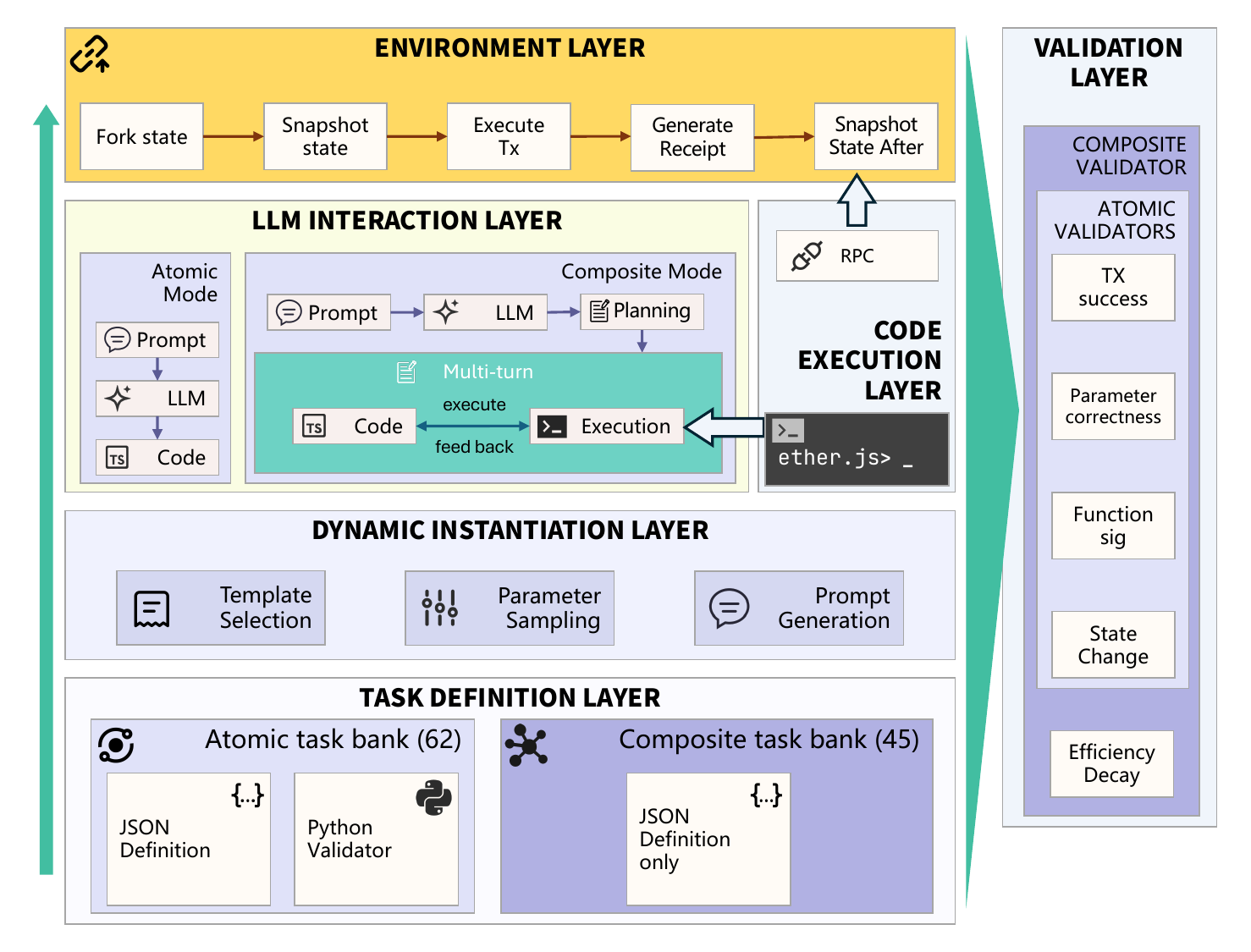}
  \caption{EVM-QuestBench evaluation architecture and end-to-end pipeline. A natural language instruction is sampled from a template pool with dynamic numeric parameters, passed to the LLM for TypeScript script generation, executed on a snapshot-isolated forked chain, and scored by task-specific validators against post-state constraints. Composite tasks additionally apply step-efficiency decay.}
  \label{fig:sec3-architecture}
\end{figure*}

Figure~\ref{fig:sec3-architecture} illustrates the layered architecture. The design is modular: adding a new atomic task requires only a JSON specification and a validator; adding a composite task requires only a JSON file. At evaluation time, the system flows through dynamic instantiation (template and parameter sampling), LLM interaction (code generation), code execution (sandboxed TypeScript runtime), transaction execution (signing and broadcasting on the fork), and validation (weighted scoring against dynamically sampled parameters).

The benchmark targets transaction script generation rather than contract synthesis. Failures commonly arise from incorrect calldata construction, unit conversion, or missing protocol prerequisites.

\subsection{Task Specification and Data Format}
Each task is stored as a JSON specification containing an identifier, metadata, natural language templates, parameter definitions, and a validation configuration. Atomic and composite tasks share a common surface structure (templates, parameters, and validation). Composite tasks additionally include an explicit workflow structure and a scoring strategy, enabling evaluation of multi transaction dependencies.

\paragraph{Atomic task schema.}
An atomic task defines a single on chain action and its expected effects. The specification includes: (i) natural language templates that render user instructions, (ii) typed parameters with default ranges, and (iii) a validator class with task specific arguments (e.g., token address, recipient address, amount, decimals). At evaluation time, the runner instantiates the validator and computes a weighted score from post execution checks.

\paragraph{Composite task schema.}
A composite task defines a multi transaction workflow and an end state condition. Each composite specification includes a \texttt{composite\_structure} field that names a workflow pattern, sets \texttt{optimal\_steps}, and enumerates step level atomic operations. Composite validation checks whether the final on chain condition holds, then applies a step efficiency decay described in Section~\ref{sec:setup}. This design prioritizes end to end completion over intermediate surface matching.

\paragraph{Running example.}
A typical instance provides an instruction such as \emph{``Swap 0.1 BNB to USDT''} together with an execution context (RPC endpoint, agent address, and a contract address map). The model returns a module whose \texttt{executeSkill} function emits a transaction request that calls the router with a concrete \texttt{to} address and \texttt{data} calldata. The validator then checks transaction success and verifies the expected balance change within the task tolerance (Section~\ref{sec:benchmark}).

\subsection{Benchmark Composition}
EVM-QuestBench contains 107 tasks (62 atomic, 45 composite) with a maximum total score of 10{,}700. Figure~\ref{fig:task-split} (Appendix~\ref{app:figures}) shows the split sizes and Figure~\ref{fig:workflow-complexity} (Appendix~\ref{app:figures}) shows the composite workflow complexity.

\textbf{Atomic tasks} require one on-chain action. The task bank is organized into three categories: \texttt{basic\_transactions} (40), \texttt{defi\_operations} (19), and \texttt{advanced\_features} (3). Basic transactions cover wallet and token operations (queries, native transfers, ERC20 transfers, approvals). DeFi operations cover swaps, liquidity, and staking. Advanced features cover edge cases (fallback handling, \texttt{delegatecall}, flashloans). Atomic tasks stress parameter correctness, unit handling, and precise state change targets.

\textbf{Composite tasks} require multi-transaction workflows with prerequisite approvals, protocol-imposed intermediate steps, and consistent parameter propagation. Examples include approve$\rightarrow$swap, approve$\rightarrow$add liquidity, and add liquidity$\rightarrow$stake. Workflows range from $2$ to $6$ optimal steps (mean $3.27$, median $3$), concentrated at $3$ steps ($53.3\%$). This split separates single-action precision from multi-step workflow completion.

\subsection{Difficulty and Coverage}
Each task is annotated with a difficulty label from \{\texttt{easy}, \texttt{easy-medium}, \texttt{medium}, \texttt{hard}\}. Atomic tasks are dominated by \texttt{medium} (56.5\%), while composite tasks skew harder (\texttt{hard}: 24.4\%), reflecting additional constraints from prerequisite handling and step ordering. Atomic tasks span subcategories including ERC20 operations, native transfers, NFT operations, swaps, staking, and queries. Composite tasks diversify by workflow motifs: batch operations, swap-centric workflows, liquidity workflows, staking workflows, and query-and-verify patterns.

\subsection{Instruction Templates and Parameterization}
Each task provides multiple natural language templates (atomic: $3$--$5$, mean $3.97$; composite: $2$--$4$, mean $2.82$), generated by an LLM to ensure diverse and realistic phrasing. At evaluation time, the runner randomly selects one template and samples numeric parameter values uniformly within predefined ranges, injecting them into the selected template and passing the resolved parameters to validators. Dynamic sampling prevents trivial memorization, forces models to generalize across arbitrary values and unit conversions, and keeps per-run token consumption low by reusing the same 107 tasks with fresh parameters across rounds.

To control for wording bias, all 373 templates were rated on a 1--5 clarity scale by three SOTA models and averaged: 336 are \textit{precise} (90.1\%) and none are \textit{vague}, confirming the pool is consistently unambiguous. Full details and the difficulty selection API are in Appendix~\ref{app:nl_difficulty}.

\subsection{Validators and Post Execution Constraints}
EVM-QuestBench uses validator-based scoring rather than reference code matching. Validators receive the same dynamically sampled parameters injected into the natural language instruction and verify that the model's output produces the expected on-chain effects for those specific values.

Each atomic task type has a dedicated validator that converts human-readable parameters to chain-native units (e.g., $0.1$ tokens $\rightarrow$ $0.1 \times 10^{18}$ wei) and compares against actual on-chain state changes. A typical atomic validator computes a weighted score from four binary checks: transaction success (30 pts), contract address correctness (20 pts), function signature correctness (20 pts), and state change verification (30 pts). Tolerance thresholds are calibrated per task family: transfer validators apply a $0.1$\% relative tolerance, approval validators require an exact match, and swap/liquidity operations use a $5$\% slippage tolerance to account for AMM non-determinism. Composite validators score the final end-state condition and apply step efficiency decay (Section~\ref{sec:setup}), reusing the corresponding atomic validator for consistency.

\subsection{Score Reporting}
We report three aggregate scores per model: Atomic score (sum over 62 tasks), Composite score (sum over 45 tasks), and Total score (sum over 107 tasks). We also report per task averages by normalizing each split sum by its task count. For summary statistics, we use a pass threshold of 60 points per task.

%% file: tex/04_setup.tex
\section{Evaluation Setup}
\label{sec:setup}

We treat correctness as an end-to-end property: a submission must construct valid transaction request objects, execute successfully in a forked environment, and satisfy post-state constraints checked by validators.

\begin{figure*}[t]
  \centering
  \includegraphics[width=\textwidth]{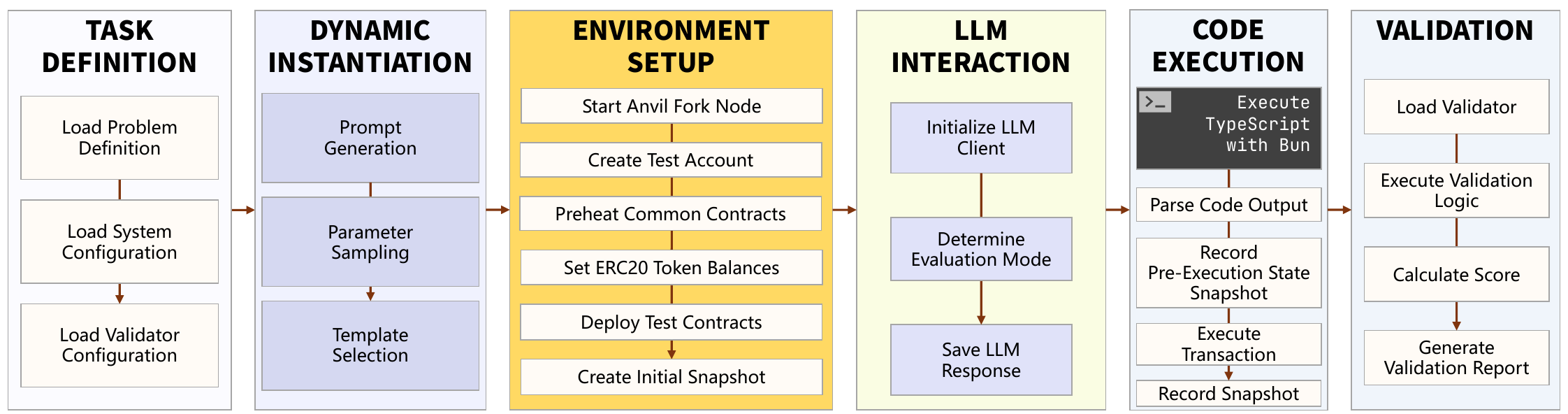}
  \caption{End-to-end evaluation pipeline of EVM-QuestBench. A natural language instruction is instantiated from a template with dynamically sampled parameters, fed to the LLM for TypeScript script generation, executed by the runner on a snapshot-isolated forked chain, and finally scored by task-specific validators against post-execution on-chain state.}
  \label{fig:eval-pipeline}
\end{figure*}

\subsection{Task Formalization}

\paragraph{Atomic tasks.}
Let $\mathcal{A} = \{a_1, a_2, \ldots, a_n\}$ denote the set of atomic tasks. Each atomic task $a_i$ is defined as a tuple:
\[
a_i = (T_i, P_i, V_i, S_i^{\max})
\]
where $T_i$ is the set of natural language templates (one is randomly selected at test time), $P_i$ is the parameter space (numeric values are sampled uniformly within predefined intervals), $V_i$ is the validator that checks post-execution state changes, and $S_i^{\max} = 100$ is the maximum score.

\paragraph{Composite tasks.}
A composite task $C$ is constructed from an ordered sequence of atomic operations:
\[
C = \bigl(a_{i_1} \rightarrow a_{i_2} \rightarrow \cdots \rightarrow a_{i_m},\ K^{\mathrm{opt}},\ V_C\bigr)
\]
where $a_{i_j} \in \mathcal{A}$ is the atomic operation at step $j$, $K^{\mathrm{opt}}$ is the optimal number of steps specified by the task definition, and $V_C$ is the composite validator that checks the final state condition. The arrow notation $\rightarrow$ indicates execution order and potential data dependencies (e.g., an approval must precede a swap that requires that allowance).

\subsection{Execution Environment}
Evaluation runs on an Anvil fork of BSC mainnet (chain ID 56), an EVM-compatible chain. We use BSC as the instantiation; the benchmark design generalizes to other EVM chains. The runner specifies the upstream RPC endpoint. The fork block height is not pinned; each evaluation run forks at the latest block. Since all 107 tasks within a run share the same fork origin and snapshot isolation (Section~\ref{subsec:isolation}) prevents cross-task state leakage, intra-run comparability is fully preserved.

Each run creates a fresh test account. The private key is generated at runtime and is never exposed to the model. Transaction signing happens in the evaluation process, separating generation from authorization and preventing the model from directly controlling keys.

Before tasks start, the account is funded with 100 BNB. The environment provisions task assets, including ERC20 tokens, LP tokens, and NFT holdings. The runner also deploys a small set of auxiliary contracts at runtime for testing.

\paragraph{Client library backend.}
All experiments use \textbf{ethers.js v6}, the de-facto standard for EVM client scripting, chosen for its broad representation in LLM training corpora and as the most equitable cross-model baseline. A \textbf{viem v2} backend is also supported; switching requires only a different system configuration file with no changes to validators or scoring. Full backend details are in Appendix~\ref{app:reproducibility}.

\subsection{Isolation}
\label{subsec:isolation}
After environment initialization, we create a snapshot of the forked chain state. Before executing each task, the runner restores the snapshot, yielding a consistent initial state per task. This prevents cross-task interference and makes task scores comparable under identical starting conditions.

\subsection{Inference and Subtask Planning}
Unless otherwise specified, we use single-shot generation for atomic tasks: each model is called once per task instance to produce a complete TypeScript module. We set the decoding temperature to $0.7$.

For composite tasks, we employ a multi-turn interaction protocol with explicit subtask planning. The LLM first enters a \textbf{planning phase}, where it analyzes the natural language instruction and decomposes the task into an ordered sequence of subtasks. Formally, the model outputs a plan:
\[
\Pi = \langle \pi_1, \pi_2, \ldots, \pi_k \rangle
\]
where each subtask $\pi_j = (t_j, c_j, p_j)$ specifies an action type $t_j$ (e.g., \texttt{approve}, \texttt{swap}, \texttt{stake}, \texttt{query}), target contract $c_j$, and parameters $p_j$. The planning objective is to minimize the total number of steps while satisfying task constraints:
\[
\min |\Pi| \quad \text{subject to} \quad \Pi \models C
\]
where $\Pi \models C$ denotes that executing $\Pi$ achieves the goal state specified by composite task $C$.

After planning, the LLM enters the \textbf{execution phase}, iterating through the planned subtasks. For each subtask, the model generates a TypeScript code block that returns a transaction request object. The runner executes the transaction on the forked chain and returns the result (success/failure, receipt, state changes) to the LLM. The model can also perform \texttt{query} actions to inspect on-chain state (balances, allowances) before executing transactions. This closed-loop interaction continues until the model signals task completion or exhausts the round budget.

\subsection{Scoring}
Each task has a maximum score of $S^{\max} = 100$.

\paragraph{Atomic scoring.}
Atomic tasks use task-specific validators. Each validator computes a weighted score from a set of post-execution checks $\mathcal{C}$:
\[
S_{\mathrm{atomic}} = \sum_{c \in \mathcal{C}} w_c \cdot \mathbf{1}\bigl[f_c(s_{\mathrm{pre}}, s_{\mathrm{post}})\bigr]
\]
where $w_c$ is the weight assigned to check $c$, $f_c$ is the check function comparing pre-execution state $s_{\mathrm{pre}}$ and post-execution state $s_{\mathrm{post}}$, and $\mathbf{1}[\cdot]$ is the indicator function. The weights satisfy $\sum_{c \in \mathcal{C}} w_c = 100$. Typical checks include transaction success (30 points), target address correctness (20 points), function signature correctness (20 points), and state-change verification through balance or allowance deltas (30 points). Where required by on-chain mechanics, validators apply tolerances to account for rounding or protocol-side price impact.

\paragraph{Composite scoring.}
Composite tasks use outcome-based scoring with step-efficiency decay. Let $K^{\mathrm{act}}$ denote the actual number of execution rounds (including failed attempts and retries up to the retry budget). The final score is:
\[
S = S_{\mathrm{base}} \cdot \min\!\left(1,\ \frac{K^{\mathrm{opt}}}{K^{\mathrm{act}}}\right)
\]
where $S_{\mathrm{base}} \in \{0, 100\}$ depends on whether the final state condition holds. This formulation rewards efficient execution: if $K^{\mathrm{act}} \leq K^{\mathrm{opt}}$, the model receives full credit; if $K^{\mathrm{act}} > K^{\mathrm{opt}}$, the score decays proportionally to the ratio of optimal to actual steps.

\paragraph{Aggregate scoring.}
Let $\mathcal{A}_{\mathrm{bench}}$ denote the set of 62 atomic tasks and $\mathcal{C}_{\mathrm{bench}}$ denote the set of 45 composite tasks. We report:
\begin{align*}
S_{\mathrm{Atomic}} &= \sum_{a \in \mathcal{A}_{\mathrm{bench}}} S_a, \\
S_{\mathrm{Composite}} &= \sum_{c \in \mathcal{C}_{\mathrm{bench}}} S_c, \\
S_{\mathrm{Total}} &= S_{\mathrm{Atomic}} + S_{\mathrm{Composite}}
\end{align*}
The maximum possible scores are $S_{\mathrm{Atomic}}^{\max} = 6{,}200$, $S_{\mathrm{Composite}}^{\max} = 4{,}500$, and $S_{\mathrm{Total}}^{\max} = 10{,}700$.

\subsection{Artifacts}
We release the benchmark codebase, task definitions, validators, and the runner. We release aggregate model scores. We do not release full per-task outputs, JSON logs, or execution traces at this time.

%% file: tex/05_results.tex
\section{Results and Analysis}
\label{sec:results}

\subsection{Leaderboard}
We evaluate 20 models on EVM-QuestBench with 5 independent evaluation rounds per model (totaling $5 \times 107 \times 20 = 10{,}700$ task executions). Each round uses freshly sampled numeric parameters, independent LLM calls (temperature $= 0.7$), and clean snapshot-isolated environments. Table~\ref{tab:leaderboard} reports the mean Atomic, Composite, and Total scores across 5 rounds, along with standard deviation (SD), observed range (min, max) over 5 rounds, and coefficient of variation (CV\%).

The top three models exceed $7{,}700$ mean total points. Claude-Sonnet-4.5 achieves the highest mean total ($8{,}236$) with low variance (CV $= 2.1\%$). 14 out of 20 models exhibit CV\% below $9\%$, demonstrating that evaluation scores are robust to the combined variance from dynamic parameter sampling, temperature-based generation, and RPC execution.

\begin{table*}[t]
\centering
\small
\setlength{\tabcolsep}{4.5pt}
\renewcommand{\arraystretch}{1.2}
\begin{tabular}{@{}rlrrrrrrcrrrr@{}}
\toprule
\textbf{\#} & \textbf{Model} & \textbf{Atom} & \textbf{Comp} & \textbf{Total} & \textbf{SD} & \textbf{CV\%} & \textbf{Range [min, max]} & & \textbf{Pass$_\text{a}$} & \textbf{Pass$_\text{c}$} & \textbf{$\bar{K}$} & \textbf{Eff\%} \\
\cmidrule(lr){1-8}\cmidrule(lr){10-13}
1  & Claude-Sonnet-4.5  & 4278 & 3958 & 8236 & 174 & 2.1 & [8017, 8403] && 35.0 & 41.0 & 3.8 & 88.0 \\
2  & Gemini-3-Pro       & 4302 & 3561 & 7863 & 267 & 3.4 & [7487, 8195] && 36.6 & 37.6 & 4.3 & 79.7 \\
3  & GPT-5              & 4128 & 3646 & 7774 & 407 & 5.2 & [7241, 8121] && 31.6 & 38.0 & 3.6 & 83.8 \\
4  & GPT-5.1            & 3715 & 3617 & 7332 & 144 & 2.0 & [7199, 7542] && 28.0 & 38.0 & 4.1 & 80.4 \\
5  & Kimi-K2-Thinking   & 3605 & 3633 & 7238 & 207 & 2.9 & [6928, 7498] && 29.2 & 33.6 & 3.6 & 84.2 \\
6  & Gemini-2.5-Flash   & 3265 & 3768 & 7033 & 127 & 1.8 & [6837, 7176] && 22.0 & 38.6 & 4.1 & 83.7 \\
\midrule
7  & GPT-OSS-120B       & 3458 & 3468 & 6926 & 436 & 6.3 & [6430, 7633] && 26.0 & 35.2 & 3.9 & 82.6 \\
8  & DeepSeek-V3.2      & 3071 & 3638 & 6709 & 384 & 5.7 & [6074, 7088] && 21.6 & 38.2 & 3.5 & 84.5 \\
9  & Grok-4-Fast        & 2980 & 3207 & 6187 & 224 & 3.6 & [5913, 6384] && 20.8 & 32.0 & 4.4 & 76.4 \\
10 & Qwen3-235B         & 3609 & 2116 & 5724 & 630 & 11.0 & [4735, 6300] && 28.0 & 11.0 & 5.8 & 62.6 \\
11 & GPT-5.1-Mini       & 2510 & 3190 & 5700 & 510 & 8.9 & [5139, 6298] && 19.4 & 31.0 & 4.7 & 73.6 \\
12 & Qwen3-Max          & 2775 & 2883 & 5658 & 229 & 4.1 & [5388, 5956] && 18.8 & 25.8 & 4.9 & 71.5 \\
13 & Claude-Haiku-4.5   & 3221 & 2367 & 5588 & 316 & 5.7 & [5208, 5899] && 24.8 & 13.0 & 5.9 & 60.7 \\
\midrule
14 & Mimo-V2-Flash      & 3189 & 1723 & 4912 & 218 & 4.4 & [4659, 5160] && 25.0 &  9.0 & 6.2 & 56.9 \\
15 & GLM-4.6            & 2881 & 1971 & 4852 & 295 & 6.1 & [4522, 5289] && 21.2 & 10.2 & 6.0 & 58.3 \\
16 & Qwen3-30B          & 1554 & 2011 & 3565 & 398 & 11.2 & [3212, 4214] && 10.8 &  8.4 & 6.1 & 58.8 \\
17 & Devstral-2512      & 2892 &   21 & 2914 & 303 & 10.4 & [2460, 3267] && 20.6 &  0.0 & 6.8 & 50.8 \\
18 & Qwen3-Coder        & 1801 &  160 & 1961 & 291 & 14.8 & [1600, 2265] && 13.2 &  0.4 & 6.7 & 53.3 \\
19 & Qwen3-Coder-Flash  &  223 &   38 &  260 & 212 & 81.4 & [40, 538]   &&  1.6 &  0.2 & 6.8 & 51.1 \\
20 & Qwen3-Coder-30B    &  140 &    5 &  145 &  61 & 42.0 & [70, 240]   &&  0.6 &  0.0 & 6.8 & 50.6 \\
\bottomrule
\end{tabular}
\caption{Leaderboard and execution quality (avg@5). Left block: Atom = Atomic (62 tasks, max 6{,}200), Comp = Composite (45 tasks, max 4{,}500), Total = 107 tasks (max 10{,}700), SD and CV\% are over Total. SD and CV\% are over Total; Range = [min, max] over 5 rounds. Spearman's $\rho = 0.960$ avg across round pairs (all $p < 0.001$, range: 0.940--0.983). Right block: Pass$_\text{a}$ = atomic tasks passed (of 62), Pass$_\text{c}$ = composite tasks passed (of 45), $\bar{K}$ = mean $K_{\text{act}}$, Eff\% = step efficiency. Pass is defined as \texttt{validation\_passed=True} (all checks satisfied).}
\label{tab:leaderboard}
\end{table*}

\subsection{Atomic versus Composite Capability Separation}
Figure~\ref{fig:atomic-composite} (Appendix~\ref{app:figures}) visualizes the Atomic--Composite plane (a full ranking bar chart is in Appendix~\ref{app:additional_analysis}, Figure~\ref{fig:app_total_ranking}). Models spread widely, indicating that single-step correctness and multi-step completion are only partially coupled.

We observe two characteristic asymmetries. Workflow-oriented models (DeepSeek-V3.2, Gemini-2.5-Flash, GPT-5.1-Mini) achieve high Composite despite weaker Atomic, suggesting stronger sequencing and end-state targeting. Precision-oriented models (Claude-Haiku-4.5, Devstral-2512) achieve high Atomic but lag on Composite, consistent with weaknesses in multi-step dependency tracking. Several code-specialized models (Qwen3-Coder-30B, Devstral-2512, Qwen3-Coder) score near zero on Composite due to repeated interface failures in multi-step workflows.

\subsection{Composite Execution Quality}
Table~\ref{tab:leaderboard} (right block) reports fine-grained composite metrics averaged over 5 rounds. We analyze step efficiency ($\bar{K}_{\text{act}}$, Eff\%) as primary dimensions.

\paragraph{Step efficiency reflects planning quality.}
Claude-Sonnet-4.5 leads with 88.0\% efficiency and the highest composite score (3{,}958 avg), completing an average of 41 out of 45 tasks. DeepSeek-V3.2 achieves the lowest $\bar{K}_{\text{act}}$ (3.5) with 84.5\% efficiency, indicating consistent per-task optimization. Lower-tier models average $\bar{K}_{\text{act}} > 5.9$, which compounds both failure risk and score decay.

\paragraph{Pass count separates tiers.}
Top models pass 33--41 of 45 composite tasks on average across 5 rounds. Claude-Haiku-4.5 passes only 13 tasks despite reasonable atomic performance, confirming that multi-step workflow completion requires capabilities beyond single-action precision. Four models (Devstral-2512, Qwen3-Coder-30B, Qwen3-Coder-Flash, Qwen3-Coder) pass fewer than 1 task on average, reflecting systematic interface or planning failures in multi-step workflows.

\paragraph{Interface compliance.}
All 20 models produce syntactically valid TypeScript modules that the runner can execute on atomic tasks. However, three code-specialized models (Qwen3-Coder-30B, Devstral-2512, Qwen3-Coder) score near zero on composite tasks due to repeated schema errors such as missing \texttt{ethers} imports or incorrect module structure in multi-step contexts. We retain these models in the leaderboard to document the full spectrum of interface-level and capability-level failures.

\subsection{Token Usage and Cost Efficiency}
\begin{figure}[t]
  \centering
  \includegraphics[width=\columnwidth]{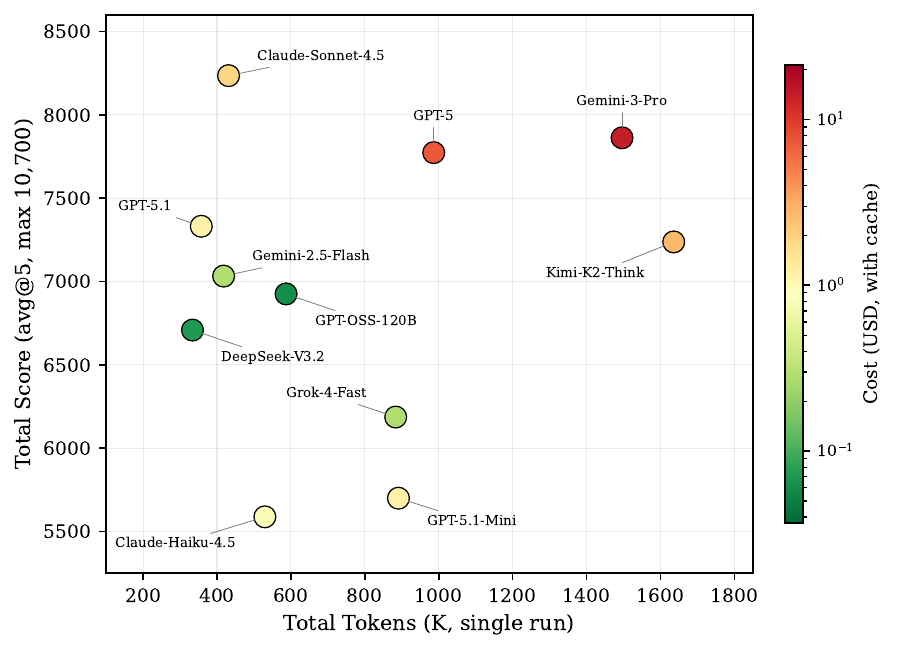}
  \caption{Total score (avg@5) versus total token usage (single run, 107 tasks). Colour encodes API cost (USD, with prompt caching). Models in the upper-left quadrant achieve high scores with low token budgets.}
  \label{fig:token-cost}
\end{figure}

Figure~\ref{fig:token-cost} plots total token consumption (single run, 107 tasks) against benchmark score, with colour encoding per-run API cost under OpenRouter cache-read pricing.

A full 107-task run costs as little as \$0.06--\$0.29 for standard models (e.g., GPT-OSS-120B at \$0.06, Gemini-2.5-Flash at \$0.29), consuming under 420K tokens. Thinking-enabled models are substantially more expensive (up to \$14.16 for Gemini-3-Pro) due to chain-of-thought token overhead, yet do not consistently outperform efficient non-thinking models. Detailed per-model cost and caching breakdowns are in Appendix~\ref{app:token_cost}. Workflow failure analysis by pattern is in Appendix~\ref{app:additional_analysis}.

%% file: tex/06_conclusion.tex
\section{Conclusion}
We presented EVM-QuestBench, an execution-grounded benchmark for natural language transaction script generation on EVM-compatible chains, instantiated on BNB Smart Chain (chain ID 56). The benchmark contains 107 tasks with Atomic and Composite splits. Model-generated TypeScript scripts are executed in a snapshot-isolated forked environment and scored by validators against post-state constraints; composite tasks additionally apply a step efficiency factor.

We evaluated 20 models under a unified runner with 5 independent rounds each. Results reveal a persistent capability gap between single-transaction precision and multi-step workflow completion. Thanks to its dynamic parameterization design, a full evaluation run requires as few as 334K tokens (as low as \$0.06), making large-scale, multi-round model comparison practically affordable.

EVM-QuestBench offers a standardized protocol for studying execution-grounded behavior in on-chain automation. We have already ported the same architecture to Solana, demonstrating the portability of the atomic/composite paradigm across heterogeneous blockchain ecosystems. Future work will expand task coverage, incorporate richer security checks for transaction intent and side effects, and evaluate LLMs' ability to generate task definitions autonomously.

%% file: tex/07_limitations.tex
\section{Limitations}
\label{sec:limitations}

\paragraph{Execution stability.}
Because EVM-QuestBench is execution-grounded, scores inherit potential instability from RPC connectivity, fork performance, and provider availability. Although this effect was negligible in our experiments, snapshot isolation and composite retries reduce but cannot fully eliminate such instability.

\paragraph{Number of evaluation rounds.}
We conduct 5 independent evaluation rounds per model and report mean scores with standard deviations and observed ranges (Table~\ref{tab:leaderboard}). While 5 rounds provide reasonable statistical power for tier-level separation, some adjacent-rank differences remain difficult to distinguish. Additional rounds would provide more stable variance estimates and enable finer-grained ranking distinctions.

\paragraph{Task coverage.}
The current benchmark contains 107 tasks, which covers a representative range of on-chain operations but remains limited in scale. We encourage community contributions to expand task coverage across protocols, chains, and operation types, and plan to maintain EVM-QuestBench as a living benchmark with ongoing community participation.

%% file: tex/A_appendix.tex
% ============================== APPENDIX (FINAL) ==============================
% Place this after \appendix in the main file, or \input{A_appendix.tex}
% Required packages in preamble:
% \usepackage{booktabs}
% \usepackage{listings}
% \usepackage{enumitem}
% \usepackage{graphicx}
% \usepackage{amsmath}

\appendix

% ==============================================================================
\section{Background: EVM Concepts and What EVM-QuestBench Measures}
\label{app:background}

This section provides a self-contained introduction to key concepts from the EVM ecosystem for readers unfamiliar with blockchain technology, and explains concretely what EVM-QuestBench is designed to evaluate.

\subsection{Core EVM Terminology}

\paragraph{Ethereum Virtual Machine (EVM).}
The EVM is a sandboxed, stack-based virtual machine that executes smart contract bytecode on a distributed network of nodes. Every EVM-compatible chain---including Ethereum mainnet, BNB Smart Chain (BSC), Polygon, and Avalanche---runs the same instruction set, so programs written for one chain can generally be ported to others with minimal changes. EVM-QuestBench is instantiated on BSC (chain ID 56) but the design generalizes to any EVM-compatible chain.

\paragraph{Smart Contract.}
A smart contract is a piece of code deployed at a fixed address on the blockchain. Once deployed, it can be called by anyone by sending a transaction that encodes the function name and its arguments as raw bytes (\emph{calldata}). The contract's logic executes deterministically on every node, and the resulting state changes (e.g., token balances, ownership records) are permanently recorded on-chain. Errors in calldata construction or parameter encoding commonly cause transactions to revert.

\paragraph{Externally Owned Account (EOA).}
An EOA is a user-controlled wallet identified by a public/private key pair. All transactions must originate from an EOA, which signs the transaction with its private key to authorize it. EVM-QuestBench creates a fresh test EOA for each evaluation run; the model generates unsigned transaction payloads that the runner signs and submits.

\paragraph{Transaction and Gas.}
A transaction is an authenticated message that transfers value or invokes a smart contract. Each transaction consumes \emph{gas}, a unit of computational cost paid by the sender in native tokens (BNB on BSC). Gas limits prevent infinite loops and prioritize efficient code. Incorrectly estimated gas limits can cause transactions to revert or fail silently.

\paragraph{BNB Smart Chain (BSC).}
BSC is an EVM-compatible blockchain operated by Binance with fast block times (${\sim}3$s) and low fees. It supports the same tooling as Ethereum. EVM-QuestBench forks BSC mainnet via Anvil to provide a realistic yet isolated on-chain environment.

\paragraph{Token Standards.}
\begin{itemize}[noitemsep,topsep=2pt]
  \item \textbf{ERC-20}: Fungible token standard. Each contract maintains a balance mapping; transfers are authorized via \texttt{approve}/\texttt{transferFrom} or direct \texttt{transfer}. Amounts must be expressed in the token's native unit (e.g., $10^{18}$ for 18-decimal tokens).
  \item \textbf{ERC-721}: Non-fungible token (NFT) standard. Each token ID is unique and owned by exactly one address.
  \item \textbf{ERC-1155}: Multi-token standard that supports both fungible and non-fungible assets in a single contract.
\end{itemize}

\paragraph{DeFi, DEX, and AMM.}
\emph{Decentralized Finance} (DeFi) refers to financial services built as smart contracts. A \emph{Decentralized Exchange} (DEX) allows token swaps without a centralized intermediary. Most DEXes use an \emph{Automated Market Maker} (AMM) model, where liquidity providers deposit token pairs into pools, and prices are determined by the ratio of reserves. PancakeSwap---the primary DEX used in EVM-QuestBench---is an AMM DEX on BSC. Swap transactions must encode the trade path, deadline, and minimum output amount; incorrect parameters cause reverts or financial loss.

\paragraph{Liquidity Provision and Staking.}
Liquidity providers deposit two tokens into an AMM pool and receive \emph{LP tokens} representing their share. LP tokens can be deposited into staking contracts to earn rewards. These workflows require multiple sequential transactions (approve $\to$ add liquidity $\to$ stake), making them natural candidates for composite task evaluation.

\subsection{What EVM-QuestBench Measures}

EVM-QuestBench evaluates whether a language model can translate a natural language instruction (e.g., ``Swap 0.1 BNB for USDT and stake the LP tokens'') into a correct, executable client-side TypeScript module. The module must:

\begin{enumerate}[noitemsep,topsep=2pt]
  \item \textbf{Correctly identify the target contract and function.} The model must select the right protocol (e.g., PancakeSwap Router vs.\ staking pool), the right function signature, and encode calldata that satisfies the ABI.
  \item \textbf{Handle chain-specific units.} Token amounts must be converted from human-readable form to on-chain representation (e.g., $0.1 \times 10^{18}$ wei for 18-decimal tokens). Swaps require slippage-tolerant minimum output values.
  \item \textbf{Satisfy protocol prerequisites.} Many operations require a prior approval transaction (ERC-20 \texttt{approve}) before the main action can execute. The model must identify and include these dependencies.
  \item \textbf{Propagate parameters across steps.} In multi-step workflows, outputs from earlier steps (e.g., LP token amounts received from liquidity addition) feed into subsequent steps (e.g., staking). The model must track and propagate these values correctly.
\end{enumerate}

The benchmark does \emph{not} evaluate contract deployment or Solidity code generation. It specifically targets the client-side scripting layer: producing unsigned transaction payloads that a standardized runner can sign and broadcast. Validators then check the resulting on-chain state---not the code itself---ensuring that evaluation measures functional correctness under execution.

% ==============================================================================
\section{Figures}
\label{app:figures}

This section contains benchmark composition and result figures referenced from the main text.

\subsection{Benchmark Composition}

\begin{figure}[h]
  \centering
  \includegraphics[width=0.42\textwidth]{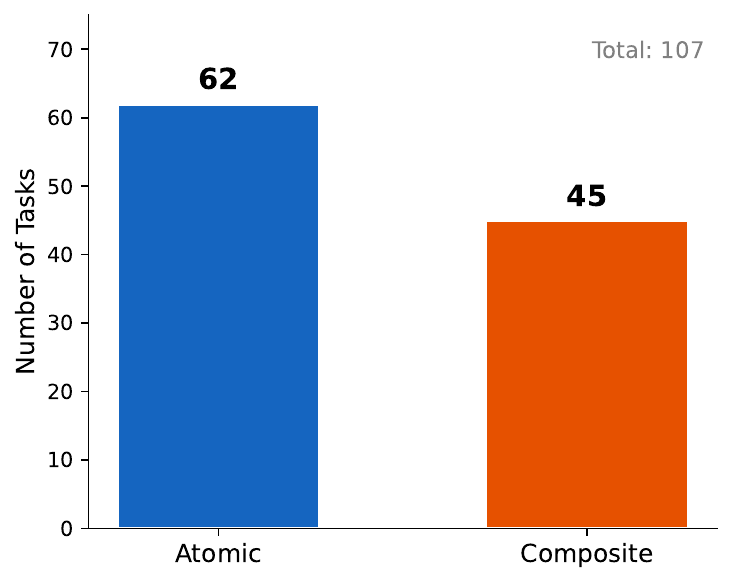}
  \caption{Task split in EVM-QuestBench: 62 atomic tasks and 45 composite tasks (107 total).}
  \label{fig:task-split}
\end{figure}

\begin{figure}[h]
  \centering
  \includegraphics[width=0.42\textwidth]{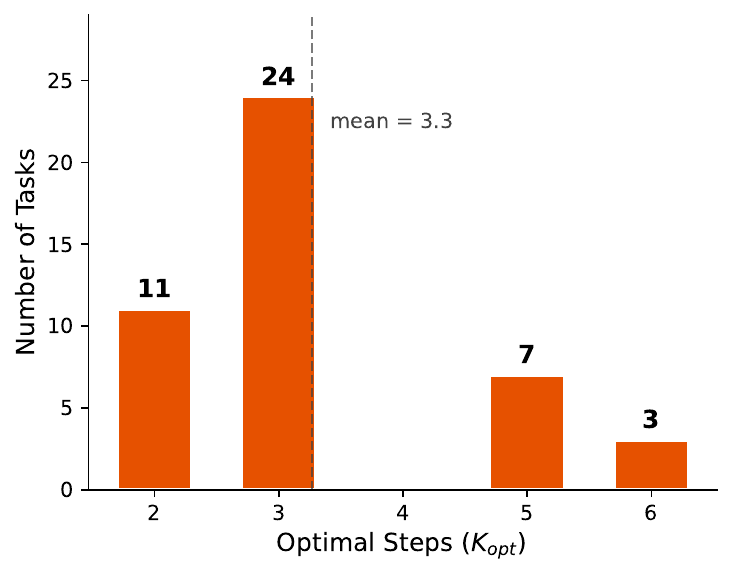}
  \caption{Distribution of optimal steps ($K_{\text{opt}}$) for composite tasks. Most workflows require 3 steps (53.3\%), with a mean of 3.3.}
  \label{fig:workflow-complexity}
\end{figure}

\subsection{Results Figures}

\begin{figure*}[h]
  \centering
  \includegraphics[width=0.85\textwidth]{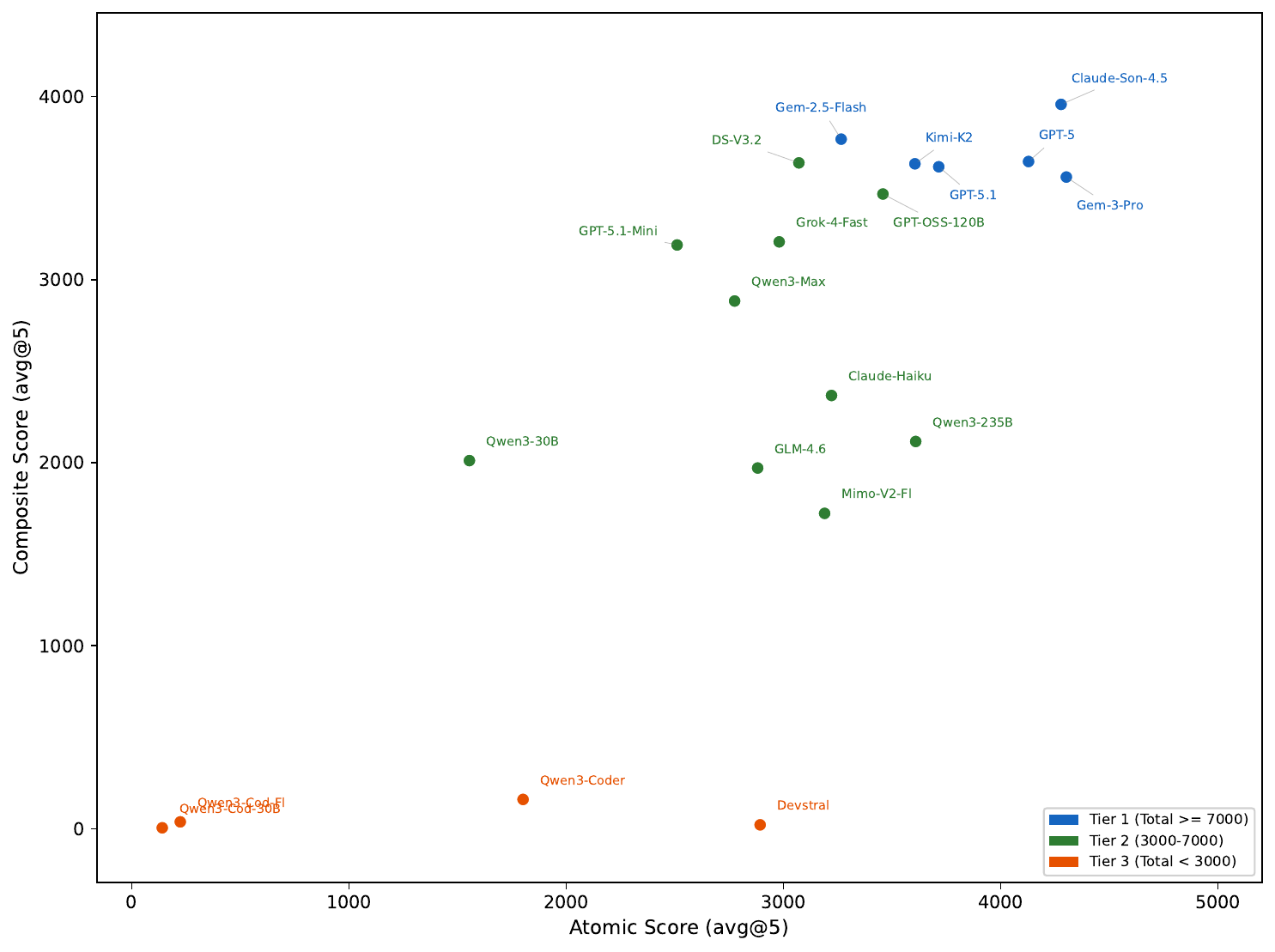}
  \caption{Atomic score versus Composite score (avg@5). Each point is a model.}
  \label{fig:atomic-composite}
\end{figure*}

\begin{figure}[h]
  \centering
  \includegraphics[width=\columnwidth]{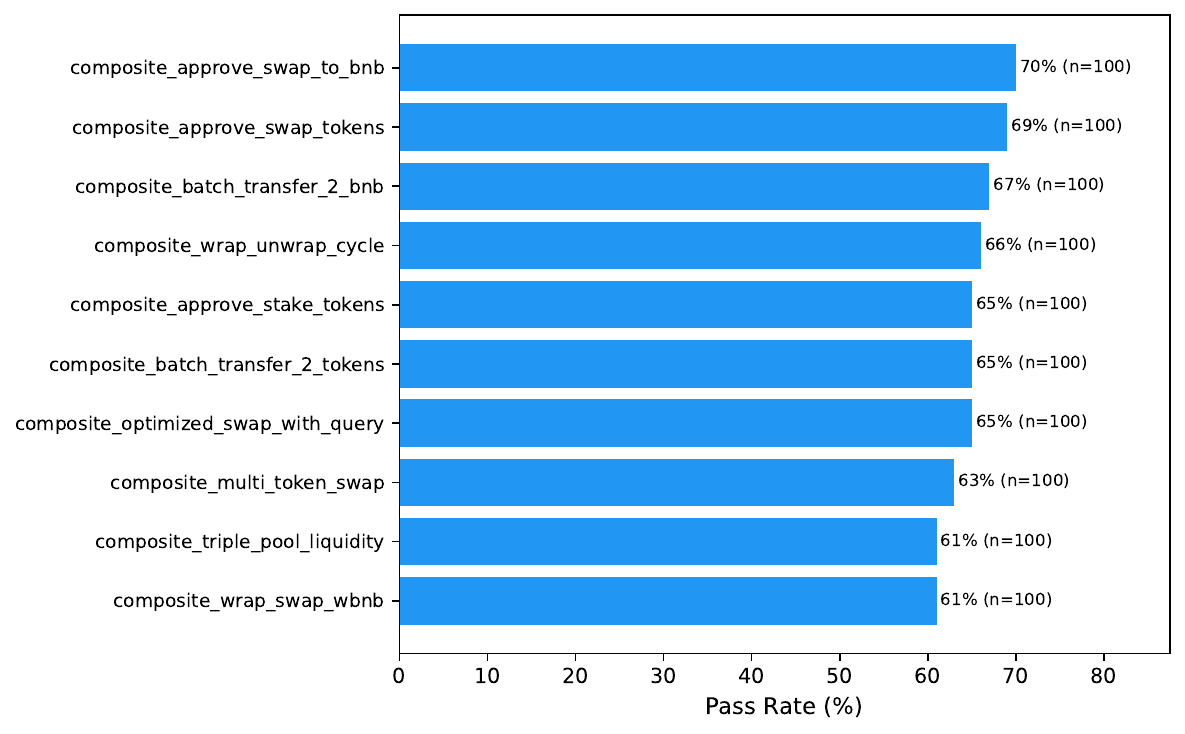}
  \caption{Composite workflow difficulty by pattern. Pass is defined as score $\ge 60$.}
  \label{fig:pattern-pass-rate}
\end{figure}

% ==============================================================================
\lstdefinelanguage{TS}{
  sensitive=true,
  morekeywords={
    async,await,const,let,var,function,return,export,import,from,
    interface,type,extends,implements,new,throw,try,catch,finally,
    Promise,Record,string,number,boolean,unknown,any,bigint
  },
  morecomment=[l]{//},
  morecomment=[s]{/*}{*/},
  morestring=[b]",
  morestring=[b]'
}

\lstset{
  basicstyle=\ttfamily\small,
  frame=single,
  columns=fullflexible,
  keepspaces=true,
  showstringspaces=false,
  breaklines=true
}

\section{Reproducibility}
\label{app:reproducibility}

This appendix summarizes the setup required to reproduce EVM-QuestBench runs and the experimental settings that affect run to run variance. For strict reproducibility, record the fork block height, model sampling parameters, and the task parameter random seed.

\subsection{Environment and Dependencies}
Experiments require a local EVM mainnet fork and a TypeScript runner.

\begin{itemize}[noitemsep,topsep=2pt]
  \item Python 3.10 or newer
  \item Node.js 18 or newer (Bun 1.0 or newer is also supported)
  \item Foundry with Anvil available in PATH
\end{itemize}

\subsection{Execution Commands}
The following commands run the benchmark for a given model identifier.

\begin{lstlisting}[language=bash]

\end{lstlisting}

\subsection{Key Experimental Settings}
Table~\ref{tab:app_repro_settings} lists the settings that most often explain score variance across reruns.

\begin{table}[t]
\centering
\footnotesize
\setlength{\tabcolsep}{4pt}
\renewcommand{\arraystretch}{1.05}
\begin{tabular}{@{}p{1.55cm}p{6.05cm}@{}}
\toprule
Item & Value \\
\midrule
Chain & EVM-compatible mainnet fork \\
Runtime & TypeScript runner using ethers.js v6 (primary; viem v2 also supported) \\
Atomic output &
One module exporting \texttt{executeSkill} that returns one
\texttt{Transaction}\texttt{Request} \\
Composite output &
Iterative calls; each round returns either one tx module or a control JSON
(query, error, submit) \\
Composite rounds &
Bounded by task \texttt{optimal\_}\texttt{steps} and a multiplier such as
\texttt{max\_rounds\_}\texttt{multiplier} \\
Non determinism &
Fork block height, model sampling parameters, random parameter sampling when seed is not fixed \\
\bottomrule
\end{tabular}
\caption{Reproducibility relevant settings.}
\label{tab:app_repro_settings}
\end{table}

\section{Prompt and Interface Specification}
\label{app:prompt_interface}

This section documents the runner interface and response validity rules used by the evaluator. The goal is to make the execution contract between model and runner explicit.

\subsection{Runner Interface}
Each model outputs a TypeScript module exporting an entry function. The runner provides \texttt{providerUrl}, the agent EOA address, and a contract address map for the local fork.

\begin{lstlisting}[language=TS]
export async function executeSkill(
  providerUrl: string,
  agentAddress: string,
  deployedContracts: Record<string, string>
): Promise<Record<string, unknown>> {
  const tx: Record<string, unknown> = {
    to: "0x...",
    data: "0x..."
  };
  return tx;
}
\end{lstlisting}

\subsection{Atomic and Composite Prompt Roles (Schematic)}
Atomic tasks require a single transaction that satisfies post execution checks. Composite tasks allow multi round interaction and apply step efficiency decay based on the number of rounds consumed.

\begin{lstlisting}
Atomic role (schematic)
Produce TypeScript in a code block.
Return exactly one transaction request object.
Use task provided parameters and addresses.

Composite role (schematic)
You may query chain state before executing.
Each round returns either one tx module or a JSON control message.
Completion is signaled with {"submit": true}.
Fewer rounds yield higher score via step efficiency decay.
\end{lstlisting}

\subsection{Schema Invalid Rules}
A response is marked \texttt{schema\_invalid} if it cannot be executed under the runner contract.

\begin{enumerate}[noitemsep,topsep=2pt]
  \item Missing exported \texttt{executeSkill}
  \item Function signature mismatch
  \item Return value is not a transaction like object
  \item Missing required \texttt{to} field
  \item Serialization failure under ethers.js
  \item No valid TypeScript code block when code is required
  \item Control JSON is not parseable in composite control rounds
\end{enumerate}

\section{Task Definition Schema}
\label{app:task_schema}

This section summarizes task fields that are most relevant for reproduction and error diagnosis.

\subsection{Atomic Task Fields}
Atomic tasks specify one on chain action and are validated by post execution constraints.

\begin{table}[t]
\centering
\scriptsize
\setlength{\tabcolsep}{3pt}
\renewcommand{\arraystretch}{1.08}
\begin{tabular}{@{}p{2.1cm}p{0.9cm}p{4.5cm}@{}}
\toprule
Field & Type & Description \\
\midrule
\texttt{id} & string & Unique task identifier \\
\texttt{category}, \texttt{subcategory} & string & Task family tags for coverage analysis \\
\texttt{difficulty} & string & easy, easy-medium, medium, hard \\
\texttt{natural\_language\_}\newline\texttt{templates} & string[] & Instruction templates for prompts \\
\texttt{parameters} & object & Typed params with sampling ranges \\
\texttt{validation} & object & Post-execution checks and weights \\
\bottomrule
\end{tabular}
\caption{Atomic task schema summary.}
\label{tab:app_atomic_schema}
\end{table}
\subsection{Composite Task Fields}
Composite tasks add a workflow template and a scoring strategy that emphasizes end to end completion.

\begin{table}[t]
\centering
\scriptsize
\setlength{\tabcolsep}{3pt}
\renewcommand{\arraystretch}{1.1}
\begin{tabular}{@{}p{2.6cm}p{0.8cm}p{4.1cm}@{}}
\toprule
Field & Type & Description \\
\midrule
\texttt{composite\_structure} & object & Workflow motif and step sequence \\
\texttt{atomic\_operations} & array & Ordered atomic steps by \texttt{atomic\_id} \\
\texttt{optimal\_steps} & int & $K_{\text{opt}}$ for step efficiency decay \\
\texttt{max\_rounds\_multiplier} & int & Upper bound on interaction rounds \\
\texttt{scoring\_strategy} & object & Validator config for end state \\
\bottomrule
\end{tabular}
\caption{Composite task schema additions.}
\label{tab:app_composite_schema}
\end{table}

% ==============================================================================
\section{NL Template Difficulty Scoring}
\label{app:nl_difficulty}

To control for potential template-wording bias, we implemented a difficulty scoring system for all 373 natural language templates. Three SOTA models (Claude Opus 4.6, GPT-5.4, and Gemini 3.1 Pro) independently rated each template on a 1--5 clarity scale (1 = maximal precision, 5 = maximal ambiguity), and ratings were averaged. Templates are classified into three tiers: \textit{precise} ($\leq 2.0$), \textit{moderate} ($2.0$--$3.5$), and \textit{vague} ($> 3.5$). Scores are stored in \texttt{nl\_template\_scores.json}.

The distribution is: 336 precise (90.1\%), 37 moderate (9.9\%), and 0 vague. This confirms that the template pool is consistently unambiguous across all task types.

At evaluation time, the \texttt{--nl-difficulty} flag supports four selection modes: \texttt{random} (default, no filtering), \texttt{precise}, \texttt{moderate}, and \texttt{vague}. All experiments in this paper use \texttt{random} mode. The difficulty API is available for controlled ablation studies on instruction clarity.

% ==============================================================================
\section{Validators and Scoring}
\label{app:validators_scoring}

EVM-QuestBench uses validator based post execution scoring rather than reference code matching. Validators check receipts and post state signals such as balance deltas, allowances, and protocol specific outcomes.

\subsection{Validator Families}
Table~\ref{tab:app_validator_families} summarizes common validator families and the constraints they evaluate.

\begin{table}[t]
\centering
\small
\setlength{\tabcolsep}{4pt}
\renewcommand{\arraystretch}{1.15}
\begin{tabular}{@{}p{0.30\columnwidth}p{0.25\columnwidth}p{0.45\columnwidth}@{}}
\toprule
Family & Typical tolerance & Examples of checks \\
\midrule
Native and ERC20 transfer & 0.1\% relative &
receipt success, recipient, amount, balance delta \\
Approval & exact match ($==$) &
spender address, allowance target \\
DEX swap & slippage tolerance &
router, path, amountIn and minOut, output delta \\
Liquidity & small relative tolerance &
token amounts, LP minted, pool state updates \\
Staking and farming & small relative tolerance &
pool identifier, staked balance delta, reward signals \\
Queries & numeric tolerance &
returned value accuracy, output format correctness \\
\bottomrule
\end{tabular}
\caption{Validator families summary.}
\label{tab:app_validator_families}
\end{table}

\subsection{Composite Step Efficiency Decay}
Composite tasks apply an outcome based base score and multiply it by a step efficiency factor.

\begin{equation}
\text{Score}_{\text{final}} = \text{Score}_{\text{base}} \times \min\left(1.0,\frac{K_{\text{opt}}}{K_{\text{act}}}\right).
\end{equation}

Here $K_{\text{opt}}$ is the task defined optimal step count and $K_{\text{act}}$ is the number of executed rounds, including retries and query rounds.

\begin{table}[t]
\centering
\small
\setlength{\tabcolsep}{10pt}
\begin{tabular}{cccc}
\toprule
$K_{\text{opt}}$ & $K_{\text{act}}$ & Decay & Base score 100 \\
\midrule
3 & 2 & 1.00 & 100 \\
3 & 3 & 1.00 & 100 \\
3 & 4 & 0.75 & 75 \\
3 & 6 & 0.50 & 50 \\
\bottomrule
\end{tabular}
\caption{Step efficiency decay examples.}
\label{tab:app_decay_examples}
\end{table}

\subsection{Validator Tolerance Sensitivity}
\label{app:tolerance_sensitivity}

To assess whether the tolerance thresholds influence evaluation outcomes, we analyzed the deviation distribution across 5 rounds $\times$ 20 models $\times$ 107 tasks (10{,}700 question instances). Table~\ref{tab:deviation_dist} reports the results.

\begin{table}[t]
\centering
\small
\setlength{\tabcolsep}{6pt}
\begin{tabular}{@{}lrr@{}}
\toprule
\textbf{Outcome} & \textbf{Count} & \textbf{\%} \\
\midrule
Score $> 0$, deviation $= 0\%$ (exact match) & 7{,}013 & 65.5\% \\
Score $= 0$, execution failure / schema error & 1{,}931 & 18.0\% \\
Score $= 0$, transaction revert               &   493 & 4.6\% \\
Score $= 0$, other failure                    & 1{,}191 & 11.1\% \\
Score $> 0$, deviation $> 0\%$ (float precision) & 72 & 0.7\% \\
\bottomrule
\end{tabular}
\caption{Actual deviation distribution across all 10{,}700 evaluated question instances (20 models $\times$ 5 rounds $\times$ 107 tasks). Non-zero deviations are all ${\sim}10^{-11}\%$ (floating-point precision), approximately $10^7\times$ smaller than the strictest tolerance.}
\label{tab:deviation_dist}
\end{table}

We varied the tolerance multiplier from $0.5\times$ to $3\times$ and re-scored all instances. Under all multiplier settings, all 20 models produce identical scores and rankings (Spearman's $\rho = 1.000$, $\Delta = 0\%$). This confirms that tolerances function as safety margins and do not influence reported scores or rankings. The bimodal nature of model outputs --- either exactly correct or catastrophically wrong (transaction revert, schema error) --- means no question instances fall in the ``close but off by a few percent'' zone where tolerance thresholds would matter.

% ==============================================================================
\section{Additional Analysis Figures}
\label{app:additional_analysis}

This section provides supporting plots used to interpret difficulty and failure modes.

\subsection{Total Score Ranking}
\begin{figure}[t]
\centering
\includegraphics[width=\linewidth]{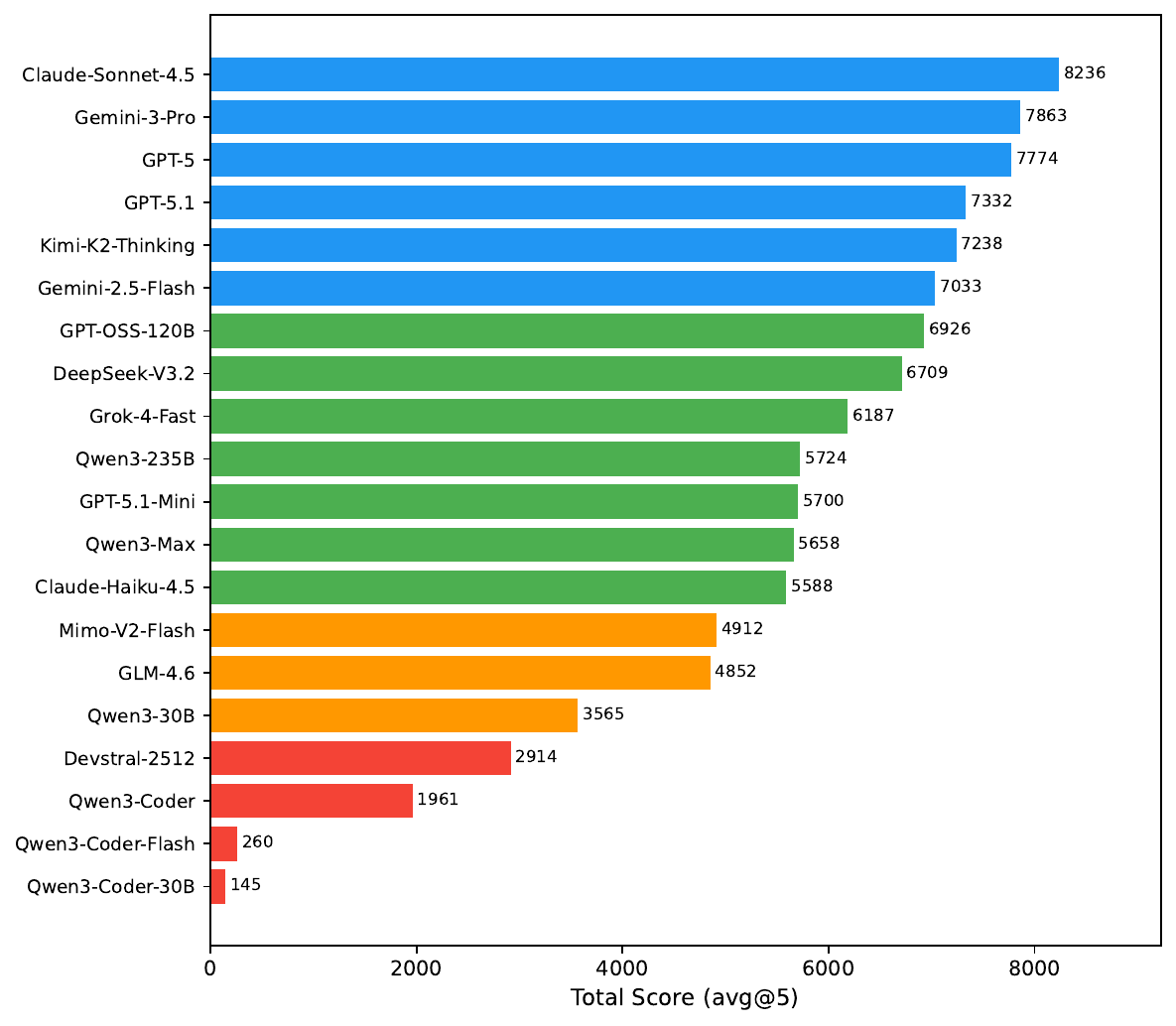}
\caption{Model ranking by total score (avg@5, 20 models).}
\label{fig:app_total_ranking}
\end{figure}

Figure~\ref{fig:app_total_ranking} shows the total score ranking across all 20 models (avg@5).

\subsection{Where Workflows Fail}
Figure~\ref{fig:app_pattern_pass_rate} reports pass rates by composite workflow pattern. The hardest patterns are multi-stage DeFi workflows combining liquidity and staking, as well as batch approval patterns. These concentrate three failure surfaces: (i) prerequisite correctness---the model must issue all required approvals before a protocol action; (ii) cross-step parameter consistency---output values from earlier steps (e.g., LP tokens received) must be correctly propagated to later steps; and (iii) execution robustness under multiple sequential transactions, where any single revert terminates the workflow and yields a zero base score.

\subsection{Composite Pattern Difficulty}
\begin{figure}[t]
\centering
\includegraphics[width=\linewidth]{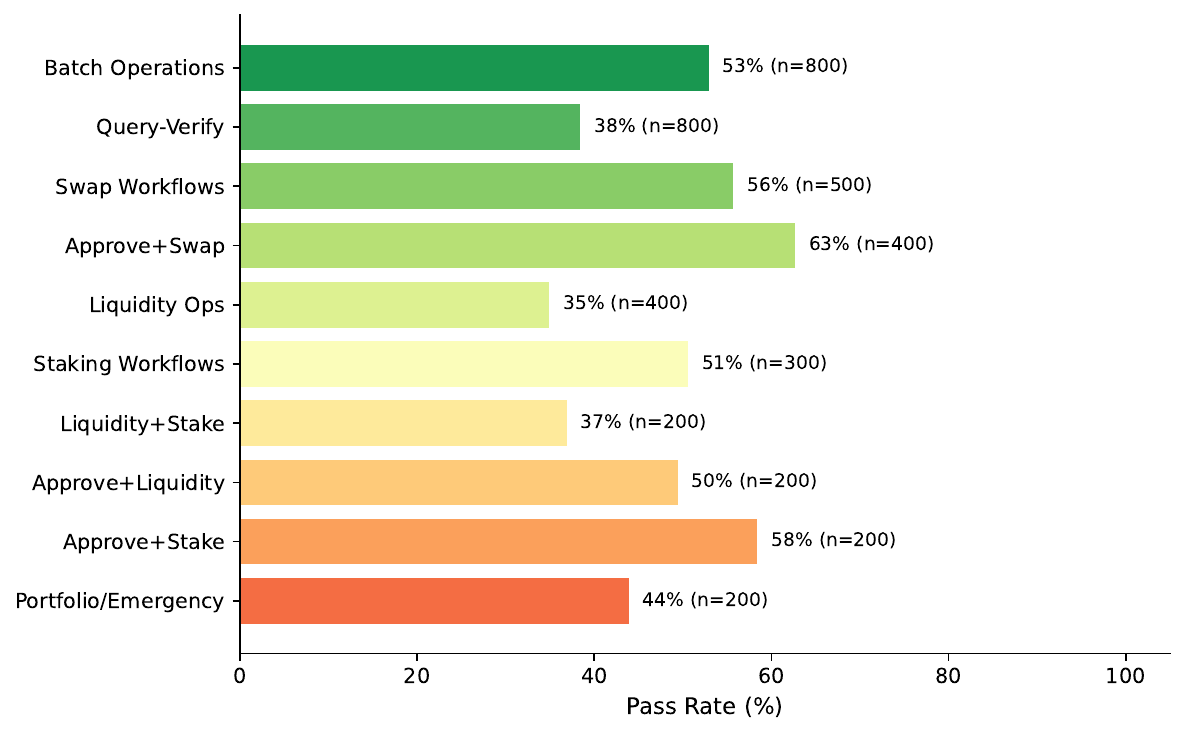}
\caption{Pass rate by composite workflow pattern for the ten most frequent patterns.}
\label{fig:app_pattern_pass_rate}
\end{figure}

Figure~\ref{fig:app_pattern_pass_rate} reports pass rates for frequent composite workflow patterns. Lower pass rates indicate higher coordination burden across steps, including prerequisite approvals and parameter propagation.

\subsection{Step Overhead Distribution}
\begin{figure}[t]
\centering
\includegraphics[width=\linewidth]{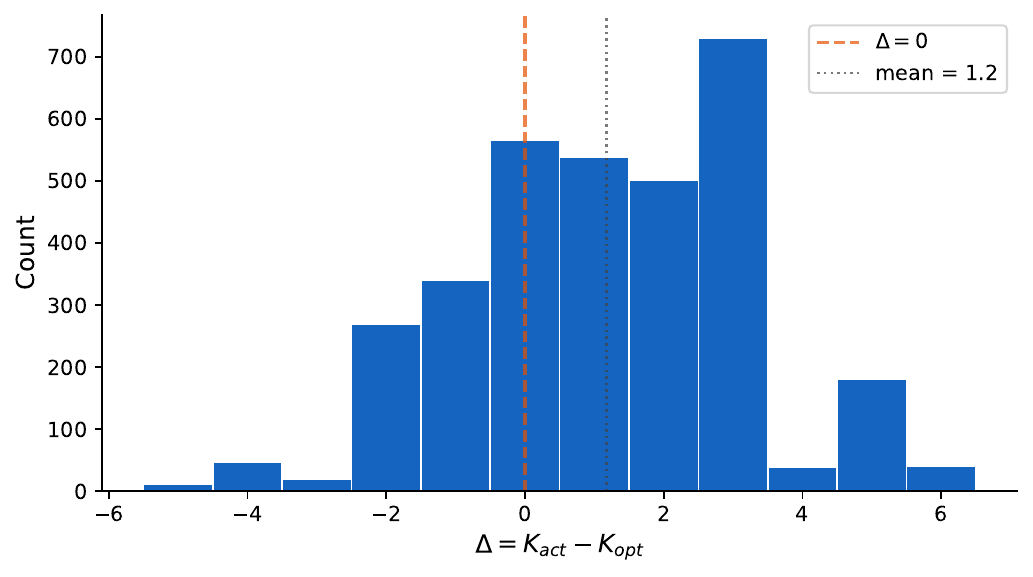}
\caption{Distribution of step overhead $\Delta = K_{\text{act}} - K_{\text{opt}}$.}
\label{fig:app_delta_steps}
\end{figure}

Figure~\ref{fig:app_delta_steps} shows the distribution of step overhead $\Delta = K_{\text{act}} - K_{\text{opt}}$ aggregated across composite runs. Positive overhead indicates extra rounds consumed by redundant queries, retries, or incorrect intermediate actions.

\subsection{Atomic Subcategory Difficulty}
\begin{figure}[t]
\centering
\includegraphics[width=\linewidth]{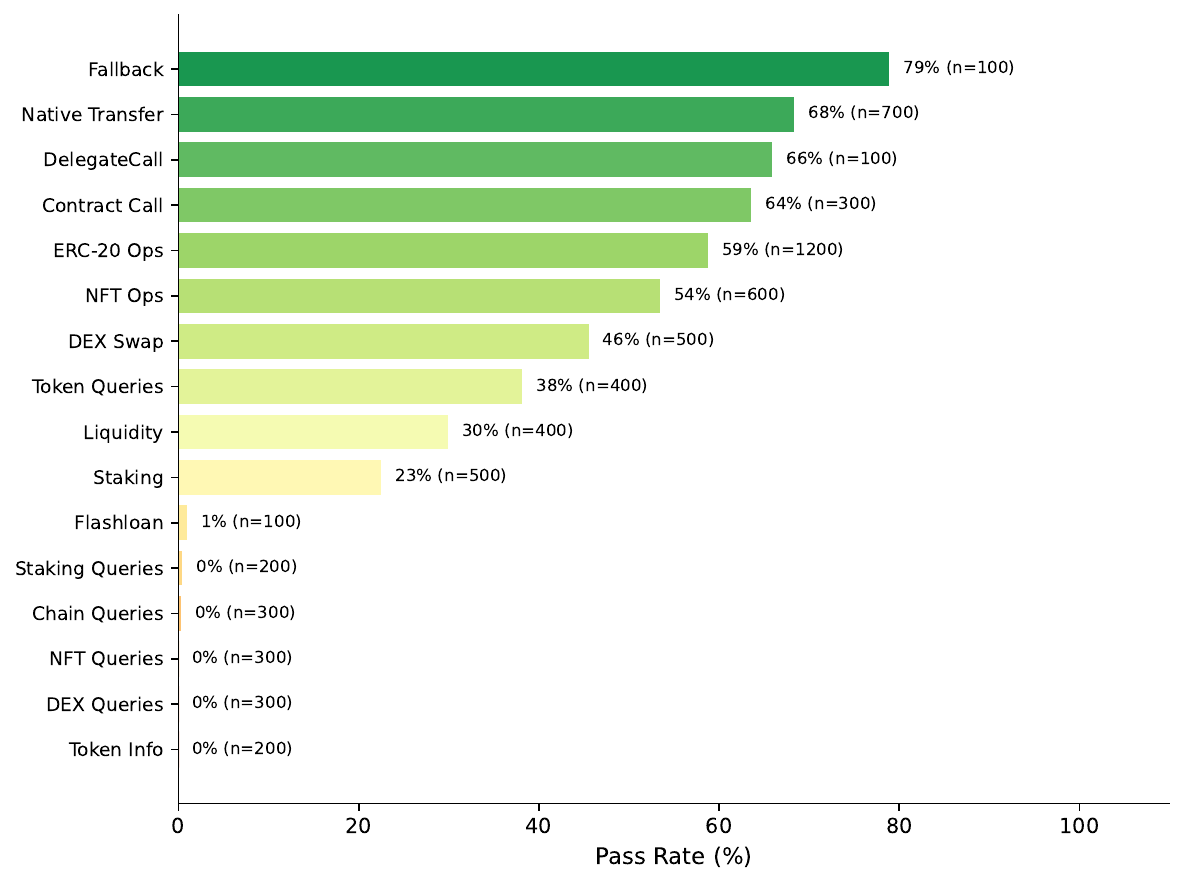}
\caption{Pass rate by atomic subcategory.}
\label{fig:app_atomic_subcat}
\end{figure}

Figure~\ref{fig:app_atomic_subcat} summarizes atomic pass rates by subcategory. This view localizes which single transaction primitives are brittle, such as query heavy tasks or protocol specific calls with strict calldata requirements.

% ==============================================================================
\section{Full Evaluation Tables}
\label{app:full_tables}

This section lists full model coverage and step overhead statistics used in the main analysis.

\subsection{Quadrant Summary}
Table~\ref{tab:app_quadrant_summary} summarizes the four capability profiles based on AtomicAvg and CompositeAvg medians (50.5 and 67.5 respectively).

\begin{table}[t]
\centering
\small
\setlength{\tabcolsep}{4pt}
\renewcommand{\arraystretch}{1.05}
\begin{tabular}{@{}p{0.26\columnwidth} r p{0.58\columnwidth}@{}}
\toprule
\textbf{Profile} & \textbf{Count} & \textbf{Interpretation} \\
\midrule
High-High   & 7 & strong precision and workflows \\
High-Low    & 3 & strong precision, weaker workflows \\
Low-High    & 3 & weaker precision, stronger workflows \\
Low-Low     & 3 & weaker on both splits \\
Atomic-Only & 4 & composite split near zero \\
\bottomrule
\end{tabular}
\caption{Quadrant summary based on AtomicAvg and CompositeAvg medians.}
\label{tab:app_quadrant_summary}
\end{table}

\subsection{Model Coverage and Quadrant Classification}
Table~\ref{tab:app_coverage_quadrant} reports coverage, normalized averages (avg@5), total score, and the quadrant label based on AtomicAvg and CompositeAvg medians.

\begin{table*}[!htbp]
\centering
\small
\setlength{\tabcolsep}{4pt}
\renewcommand{\arraystretch}{1.12}
\begin{tabular}{@{}p{0.26\textwidth}cccccc@{}}
\toprule
Model & Atomic & Composite & AtomicAvg & CompAvg & Total & Quadrant \\
\midrule
Claude-Sonnet-4.5 & 62/62 & 45/45 & 69.0 & 88.0 & 8235.8 & High-High \\
Gemini-3-Pro & 62/62 & 45/45 & 69.4 & 79.1 & 7863.0 & High-High \\
GPT-5 & 62/62 & 45/45 & 66.6 & 81.0 & 7773.7 & High-High \\
GPT-5.1 & 62/62 & 45/45 & 59.9 & 80.4 & 7331.7 & High-High \\
Kimi-K2-Thinking & 62/62 & 45/45 & 58.1 & 80.7 & 7238.3 & High-High \\
Gemini-2.5-Flash & 62/62 & 45/45 & 52.7 & 83.7 & 7033.4 & High-High \\
GPT-OSS-120B & 62/62 & 45/45 & 55.8 & 77.1 & 6925.7 & High-High \\
\midrule
Qwen3-235B & 62/62 & 45/45 & 58.2 & 47.0 & 5724.5 & High-Low \\
Claude-Haiku-4.5 & 62/62 & 45/45 & 52.0 & 52.6 & 5588.4 & High-Low \\
Mimo-V2-Flash & 62/62 & 45/45 & 51.4 & 38.3 & 4912.3 & High-Low \\
\midrule
DeepSeek-V3.2 & 62/62 & 45/45 & 49.5 & 80.9 & 6709.2 & Low-High \\
Grok-4-Fast & 62/62 & 45/45 & 48.1 & 71.3 & 6187.0 & Low-High \\
GPT-5.1-Mini & 62/62 & 45/45 & 40.5 & 70.9 & 5700.0 & Low-High \\
\midrule
Qwen3-Max & 62/62 & 45/45 & 44.8 & 64.1 & 5658.5 & Low-Low \\
GLM-4.6 & 62/62 & 45/45 & 46.5 & 43.8 & 4852.0 & Low-Low \\
Qwen3-30B & 62/62 & 45/45 & 25.1 & 44.7 & 3565.5 & Low-Low \\
\midrule
Devstral-2512 & 62/62 & 45/45 & 46.7 & 0.5 & 2913.5 & Atomic-Only \\
Qwen3-Coder & 62/62 & 45/45 & 29.0 & 3.6 & 1961.0 & Atomic-Only \\
Qwen3-Coder-Flash & 62/62 & 45/45 & 3.6 & 0.8 & 260.5 & Atomic-Only \\
Qwen3-Coder-30B & 62/62 & 45/45 & 2.3 & 0.1 & 145.0 & Atomic-Only \\
\bottomrule
\end{tabular}
\caption{Model coverage and quadrant classification (avg@5). Averages are normalized to the 0--100 scale. Medians: AtomicAvg\,=\,50.5, CompositeAvg\,=\,67.5.}
\label{tab:app_coverage_quadrant}
\end{table*}

\subsection{Step Overhead and Decay Impact Summary}
Table~\ref{tab:app_step_decay} reports overhead statistics (avg@5) for models with non-trivial composite scores. The overhead statistic is $\Delta = K_{\text{act}} - K_{\text{opt}}$.

\begin{table}[!htbp]
\centering
\footnotesize
\setlength{\tabcolsep}{3pt}
\renewcommand{\arraystretch}{1.05}
\resizebox{\columnwidth}{!}{%
\begin{tabular}{p{2.7cm}cccc}
\toprule
Model & Mean $\Delta$ & Over-opt (\%) & P90 $\Delta$ & Comp.\ score \\
\midrule
DeepSeek-V3.2 & +0.05 & 44 & 3 & 3638.4 \\
Kimi-K2-Thinking & +0.17 & 42 & 3 & 3633.3 \\
GPT-5 & +0.19 & 48 & 3 & 3645.7 \\
Claude-Sonnet-4.5 & +0.34 & 36 & 2 & 3958.0 \\
GPT-OSS-120B & +0.41 & 47 & 3 & 3468.1 \\
Gemini-2.5-Flash & +0.67 & 45 & 3 & 3768.0 \\
GPT-5.1 & +0.70 & 56 & 3 & 3617.1 \\
Gemini-3-Pro & +0.83 & 59 & 3 & 3561.0 \\
Grok-4-Fast & +1.02 & 62 & 3 & 3206.8 \\
GPT-5.1-Mini & +1.24 & 67 & 3 & 3189.8 \\
Qwen3-Max & +1.35 & 63 & 3 & 2883.5 \\
Qwen3-Coder & +2.36 & 86 & 4 & 160.0 \\
Qwen3-235B & +2.47 & 84 & 5 & 2115.9 \\
Claude-Haiku-4.5 & +2.51 & 88 & 5 & 2367.4 \\
GLM-4.6 & +2.55 & 89 & 5 & 1971.0 \\
Mimo-V2-Flash & +2.76 & 92 & 5 & 1723.1 \\
Qwen3-30B & +2.88 & 92 & 5 & 2011.5 \\
\bottomrule
\end{tabular}%
}
\caption{Step overhead and decay impact summary (avg@5). Models with near-zero composite scores (Devstral-2512, Qwen3-Coder-Flash, Qwen3-Coder-30B) are omitted.}
\label{tab:app_step_decay}
\end{table}

\section{End-to-End Walkthrough}
\label{app:e2e_walkthrough}

We illustrate the complete evaluation pipeline with one atomic task: \texttt{bnb\_transfer\_percentage}.

\paragraph{Step 1: Parameter Instantiation.}
The runner selects a natural language template and samples numeric parameters:
\begin{itemize}[noitemsep,topsep=2pt]
  \item \textbf{Template selected:} ``Transfer \{percentage\}\% of my BNB balance to \{recipient\}.''
  \item \textbf{Parameters sampled:} \texttt{percentage} $= 15$, \texttt{recipient} $=$ \texttt{0xA1b2...C3d4} (randomly generated).
  \item \textbf{Final instruction:} ``Transfer 15\% of my BNB balance to 0xA1b2...C3d4.''
\end{itemize}

\paragraph{Step 2: Script Generation.}
The model receives the instruction along with system context (chain ID, available libraries, account address, contract ABIs). It generates a TypeScript module that:
\begin{enumerate}[noitemsep,topsep=2pt]
  \item Queries the current BNB balance via \texttt{provider.getBalance(account)}.
  \item Computes \texttt{amount = balance * 15n / 100n}.
  \item Returns a transaction request: \texttt{\{to: recipient, value: amount\}}.
\end{enumerate}

\paragraph{Step 3: Execution.}
The runner signs and submits the transaction on the forked chain. The transaction executes successfully (receipt status $= 1$). The runner records pre-execution and post-execution balances for both the sender and recipient.

\paragraph{Step 4: Validation.}
The \texttt{bnb\_transfer\_percentage} validator performs four checks:
\begin{enumerate}[noitemsep,topsep=2pt]
  \item \textbf{Transaction Success} (30 pts): receipt status $= 1$. \checkmark
  \item \textbf{Recipient Correct} (20 pts): \texttt{tx.to} matches the sampled recipient. \checkmark
  \item \textbf{Transfer Amount} (20 pts): actual transferred amount within 0.1\% of expected ($\text{balance} \times 15\%$). \checkmark
  \item \textbf{Balance Change} (30 pts): sender balance decreased by amount $+$ gas; recipient balance increased by amount, both within 0.1\% tolerance. \checkmark
\end{enumerate}
\textbf{Final score:} $30 + 20 + 20 + 30 = 100$ out of 100.

\paragraph{Key design points.} The expected transfer amount is computed dynamically from the fork state (not hardcoded), so the ground truth is always consistent with the execution environment. The tolerance on balance change (0.1\%) absorbs floating-point precision differences without admitting materially incorrect transfers.

\section{Case Studies}
\label{app:case_studies}

This section provides representative workflows that illustrate common failure modes. The goal is to highlight failure points such as parameter propagation, prerequisite handling, and protocol sequencing.

\subsection{Case 1: Complex Multi step DeFi Workflow}
\textbf{Task:} \texttt{composite\_complete\_swap\_stake\_workflow}. \\
\textbf{Workflow:} query, approve, swap, add liquidity, stake, verify. \\
\textbf{Optimal steps:} 6. \quad \textbf{Pass rate:} 59\%.

This task combines prerequisite checks with multiple protocol interactions. Failures often arise from incorrect swap path or slippage, mismatched token ratios during liquidity provision, or staking to an incorrect pool. Models that succeed typically keep parameters consistent across all steps and avoid redundant query rounds that increase $K_{\text{act}}$.

\subsection{Case 2: Batch Approval Workflow}
\textbf{Task:} \texttt{composite\_batch\_approve\_2\_tokens}. \\
\textbf{Workflow:} approve token A, approve token B, execute the downstream action. \\
\textbf{Optimal steps:} 3. \quad \textbf{Pass rate:} 56\%.

This pattern stresses multi-transaction sequencing and nonce handling. Common failures include nonce conflicts under rapid submissions, gas estimation errors for back-to-back approvals, and incomplete approvals that allow only one token to proceed.

\subsection{Case 3: Natural Language Misinterpretation}
\textbf{Task:} \texttt{bnb\_transfer\_percentage}. \\
\textbf{Instruction:} ``Transfer 15\% of my BNB balance to 0xA1b2...C3d4.'' \\
\textbf{Failure mode:} The model interprets ``15\%'' as a fixed amount of 15 BNB rather than 15\% of the current balance. This produces a correct transaction that executes successfully but transfers the wrong amount --- the validator detects a deviation exceeding the 0.1\% tolerance on the Transfer Amount check. \\
\textbf{Affected models:} Observed in lower-tier models that lack numeric reasoning about percentage-based instructions.

\subsection{Case 4: Planning-Level Failure}
\textbf{Task:} \texttt{composite\_approve\_and\_swap}. \\
\textbf{Instruction:} ``Swap 0.05 BNB for USDT on PancakeSwap.'' \\
\textbf{Failure mode:} The model directly calls the PancakeSwap Router's \texttt{swapExactETHForTokens} without first checking whether the WBNB$\to$USDT path exists or setting appropriate slippage. The transaction reverts due to insufficient output amount. Some models omit the prerequisite approval step entirely when the swap requires an ERC-20 input token. \\
\textbf{Impact:} The model scores 0 on this task. Step-efficiency decay is irrelevant since the base score is 0.

\subsection{Case 5: Partial Execution in Multi-Step Workflows}
\textbf{Task:} \texttt{composite\_complete\_swap\_stake\_workflow}. \\
\textbf{Instruction:} ``Swap BNB for CAKE, add CAKE-BNB liquidity, then stake the LP tokens.'' \\
\textbf{Failure mode:} The model successfully completes the swap and liquidity addition but fails on the staking step due to an incorrect staking pool address or mismatched LP token. The validator checks the final state (LP tokens staked) and assigns a base score of 0 despite partial progress. \\
\textbf{Observation:} This pattern accounts for the majority of composite failures among mid-tier models, where individual actions succeed but end-to-end workflow orchestration breaks down.

% ==============================================================================
\section{Token Usage and Cost Analysis}
\label{app:token_cost}

Figure~\ref{fig:token-cost} (Section~\ref{sec:results}) plots total token consumption against benchmark score for a single 107-task run. This appendix provides the detailed breakdown.

\paragraph{Token-efficient models.}
Token budgets vary by over $4\times$ across the top-11 models. Claude-Sonnet-4.5 achieves the highest score (8{,}236) while consuming only 431K tokens per run, making it both the most accurate and one of the most token-efficient models. Other efficient models include DeepSeek-V3.2 (334K tokens, \$0.07), GPT-5.1 (358K, \$1.15), and Gemini-2.5-Flash (418K, \$0.29), all scoring above 6{,}700 with fewer than 420K tokens per run.

\paragraph{Thinking-enabled models.}
Thinking-enabled models allocate substantially more output tokens to chain-of-thought reasoning: Kimi-K2-Thinking (1{,}636K, \$2.74), Gemini-3-Pro (1{,}496K, \$14.16), and GPT-5 (986K, \$7.48) each exceed 900K tokens, with output tokens dominating their budgets (82--93\% of total).

\paragraph{Cost range and caching.}
Cost spans three orders of magnitude. GPT-OSS-120B achieves rank 7 at just \$0.06 per run, making it $230\times$ cheaper than Gemini-3-Pro (rank 2, \$14.16) despite only a 12\% score gap. Claude-Sonnet-4.5 incurs \$1.91 per run due to its premium output pricing (\$15/M), yet its low token count keeps it far below the thinking models in absolute cost. Prompt caching reduces input costs by up to 30\% for models with high cache-read ratios (Claude, Gemini), though output-heavy thinking models benefit minimally (2--4\% savings) since caching applies only to input tokens.